\documentclass[letterpaper, 10 pt, conference]{ieeeconf}  

\IEEEoverridecommandlockouts                              

\usepackage{graphicx}
\usepackage{subcaption} 
\usepackage{comment}
\usepackage{graphicx}
\usepackage{subcaption}
\usepackage{array}
\usepackage{caption}
\usepackage{amsmath,amsfonts} 
\usepackage{amssymb}
\usepackage{gensymb}
\title{\LARGE \bf
A Generative Adversarial Network-based Method for LiDAR-Assisted Radar Image Enhancement
}

\author{Thakshila Thilakanayake$^{1}$ Oscar De Silva$^{1}$, Thumeera R. Wanasinghe$^{2}$, George K. Mann$^{1}$, Awantha Jayasiri$^{3}$ 
\thanks{*This work was supported in part by the National Research Council of Canada’s Artificial Intelligence for Logistics Program, in part by the Natural Sciences and Engineering Research Council of Canada, and in part by the Memorial University of Newfoundland.
}
\thanks{$^{1}$Department of Mechanical and Mechatronics Engineering, Memorial University of Newfoundland, St. John’s, NL, Canada.
        {\tt\small dthilakanaya@mun.ca, oscar.desilva@mun.ca, gmann@mun.ca}}%
\thanks{$^{2}$Department of Electrical and Computer Engineering, Memorial University of Newfoundland, St. John’s, NL, Canada.
        {\tt\small thumeerawa@mun.ca}}%
\thanks{$^{3}$Aerospace Research
Centre, National Research Council Canada, Ottawa, ON, Canada.
        {\tt\small awantha.jayasiri@nrc-cnrc.gc.ca}}%
}

\begin{document}

\maketitle
\thispagestyle{empty}
\pagestyle{empty}

\begin{abstract}

This paper presents a generative adversarial network (GAN) based approach for radar image enhancement. Although radar sensors remain robust for operations under adverse weather conditions, their application in autonomous vehicles (AVs) is commonly limited by the low-resolution data they produce. The primary goal of this study is to enhance the radar images to better depict the details and features of the environment, thereby facilitating more accurate object identification in AVs. The proposed method utilizes high-resolution, two-dimensional (2D) projected light detection and ranging (LiDAR) point clouds as ground truth images and low-resolution radar images as inputs to train the GAN. The ground truth images were obtained through two main steps. First, a LiDAR point cloud map was generated by accumulating raw LiDAR scans. Then, a customized LiDAR point cloud cropping and projection method was employed to obtain 2D projected LiDAR point clouds. The inference process of the proposed method relies solely on radar images to generate an enhanced version of them. The effectiveness of the proposed method is demonstrated through both qualitative and quantitative results. These results show that the proposed method can generate enhanced images with clearer object representation compared to the input radar images, even under adverse weather conditions.

\end{abstract}

\section{INTRODUCTION}

Autonomous vehicles (AVs) are equipped with an array of sensors to sense their surroundings, estimate their pose, and plan collision-free trajectories.  The typical surrounding perception package of AVs consists of camera, light detection and ranging (LiDAR), and radar sensors \cite{n2}. 

\begin{figure*}[!h]
     \centering
     \begin{subfigure}[]{0.15\textwidth}
         \centering
         \includegraphics[width=\textwidth]{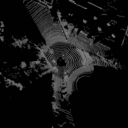}
         \caption{}
         \label{fig1rg}
     \end{subfigure}
     \hfill
     \begin{subfigure}[]{0.15\textwidth}
         \centering
         \includegraphics[width=\textwidth]{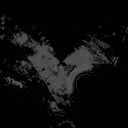}
         \caption{}
         \label{fig1rs}
     \end{subfigure}
     \hfill
     \begin{subfigure}[]{0.15\textwidth}
         \centering
         \includegraphics[width=\textwidth]{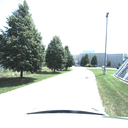}
         \caption{}
         \label{fig1lg}
     \end{subfigure}
     \hfill
     \begin{subfigure}[]{0.15\textwidth}
         \centering
         \includegraphics[width=\textwidth]{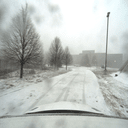}
         \caption{}
         \label{fig1ls}
     \end{subfigure}
     \hfill
     \begin{subfigure}[]{0.15\textwidth}
         \centering
         \includegraphics[width=\textwidth]{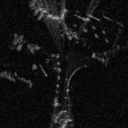}
         \caption{}
         \label{fig1cg}
     \end{subfigure}
     \hfill
    \begin{subfigure}[]{0.15\textwidth}
         \centering
         \includegraphics[width=\textwidth]{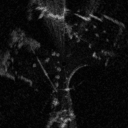}
         \caption{}
         \label{fig1cs}
     \end{subfigure}
\caption{Illustration of LiDAR, camera, and radar images of a sample location from the Boreas dataset for sunny and snowy weather conditions: (a) 2D projected, raw LIDAR scan for sunny weather conditions, (b) 2D projected, raw LIDAR scan for snowy weather conditions, (c) camera image for sunny weather conditions, (d) camera image for snowy weather conditions, (e) radar image for sunny weather conditions, (f) radar image for snowy weather conditions. }
\label{fig1}
\end{figure*}

Most of the current LiDAR sensors deliver accurate distance measurements with a large field of view and dense three-dimensional (3D) point clouds \cite{r23}. This capability makes them useful for tasks such as object detection, classification, and segmentation \cite{r46}. However, the performance of the LiDAR systems is significantly affected and the detection range is drastically reduced in adverse weather conditions such as rain, snow, mist, and fog, potentially leading to inaccurate LiDAR point clouds.  Fig \ref{fig1} displays two-dimensional (2D) projected LiDAR, camera, and radar images of a sample location collected in sunny and snowy weather conditions, sourced from the Boreas dataset \cite{r11}. It is evident from Fig \ref{fig1rg} and \ref{fig1rs} that the detection range of the LiDAR sensor has significantly decreased and there is missing information in the acquired point cloud during snowy conditions when compared to the LiDAR point cloud acquired in good weather conditions. These issues are primarily due to the weakening of the detection capability of LiDAR sensors in adverse weather conditions \cite{n1,r45}.

Camera-based perceptual systems also often struggle in adverse weather conditions due to low light, obstructed views, reflections, and glare. Adverse weather can alter object appearances, lead to misclassifications, and disrupt depth perception, making accurate identification, and segmentation challenging \cite{r15}. When Figures \ref{fig1lg} and \ref{fig1ls} are compared, they demonstrate a significant decrease in overall camera visibility during snowy weather conditions compared to sunny weather. 

In contrast, radar sensors are robust in adverse weather conditions as they generally operate at frequencies that are less affected by precipitation or reduced visibility \cite{r10}. This characteristic enhances the utility of radar sensors for AVs in adverse weather conditions. Fig \ref{fig1cg} and \ref{fig1cs} illustrate that the radar image remains largely consistent, regardless of weather conditions, demonstrating the robustness of the radar sensors in adverse weather conditions. Consequently, several studies have proposed radar-based object detection \cite{r10,r52,r53} and semantic segmentation methods \cite{r54,r55,r56}. 

Despite the growing popularity of radar based object detection and semantic segmentation methods, the low-resolution data provided by the radar sensors and the limited availability of radar-based, annotated datasets make accurate object identification-related applications challenging \cite{r33, r34, r28}. This limitation highlights the need for a method to enhance low-resolution radar images to support autonomous navigation in adverse weather conditions.

Recently, Generative Adversarial Network (GAN)-based methods have gained prominence and showcased outstanding performance, often surpassing image processing and supervised deep learning (DL) based approaches in the domain of low-quality image enhancement and image super-resolution. Ledig C. et al. \cite{n8} proposed a GAN-based method for single image super-resolution, which can be effectively used for generating photo realistic images from heavily downsampled image inputs. The findings of the study \cite{r17} demonstrate that a GAN can enhance the quality of low-resolution images captured by handheld ultrasound devices to match those taken with high-end, large-scale ultrasound machines. Furthermore, these findings highlight that GANs are capable of improving the resolution and visibility of underlying structures in low-resolution images. The Pix2Pix framework by Isola et al. \cite{n9} further extends the use of GANs into the domain of image-to-image translation, making it particularly effective for tasks that involve converting between different types of image data, such as synthesizing photos from label maps and reconstructing objects from edge maps. The findings of the studies on GAN-based image enhancement, image-to-image translation, and image super-resolution within the literature provide significant inspiration for employing GANs for radar image enhancement.

The work presented in this paper proposes a GAN-based method, designed to enhance low-resolution radar images. The proposed method utilizes high-resolution, 2D projected LiDAR point clouds as ground truth images and low-resolution radar images as inputs for training the GAN. The inference process of the proposed method relies solely on radar images to generate an enhanced version of them. The proposed method enables AVs to rely solely on radar sensors for object identification tasks, particularly when the camera and LiDAR systems are unavailable or compromised by adverse weather conditions. The main contributions of this study are,

\begin{itemize}

\item Development of a radar image enhancement method that generates enhanced images with clearer object representation compared to the input radar images.

\item A high-resolution ground truth training data generation method combining LiDAR mapping and global navigation satellite system (GNSS) measurements.

\item Validation of the enhancement performance of the proposed method using field datasets which include good and adverse weather conditions.

\end{itemize}

\begin{figure*}[h]
  \centering
  \includegraphics[scale=0.24]{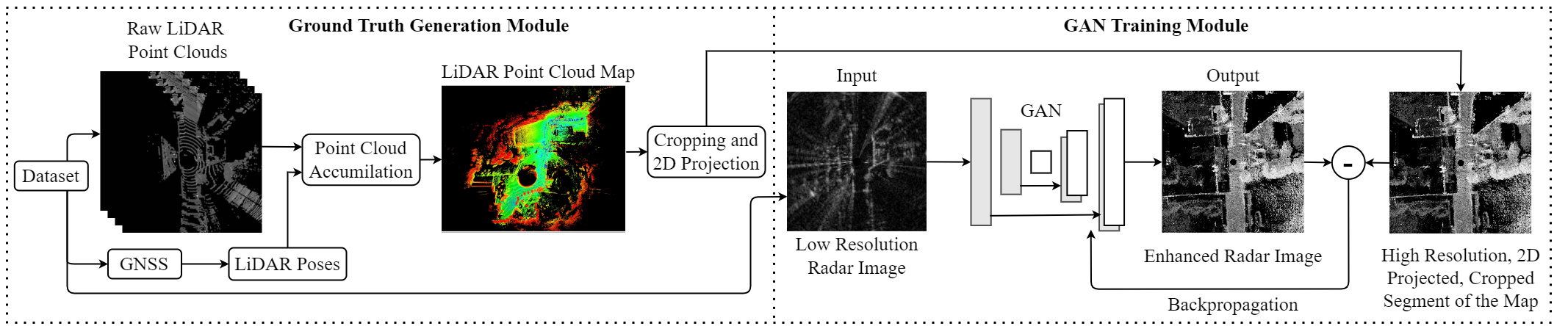}
  \caption{Illustration of the overall system diagram of the proposed radar image enhancement method}
  \label{fig2}
\end{figure*}

\section{Methodology}

\begin{figure}[h]
  \centering
  \includegraphics[scale=0.35]{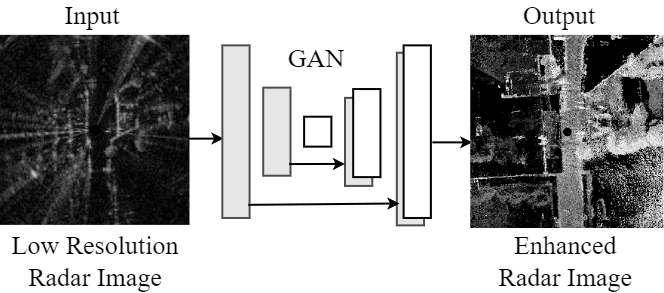}
  \caption{Illustration of the inference process of the proposed radar image enhancement method}
  \label{fig2in}
  
\end{figure}

The system overview of the proposed GAN-based radar image enhancement method is shown in Fig. \ref{fig2}. This system consists of two primary modules:  (1) the ground truth LiDAR point cloud generation module and (2) GAN training module. The former is responsible for generating ground truth images for GAN training. This involves two main steps. The first step generates a LiDAR point cloud map by accumulating raw LiDAR scans based on their poses given by the GNSS. In the second step, a customized cropping and 2D projection method is employed for cropping the LiDAR point cloud map into instances and converting the cropped instances into pixel-wise matching 2D projected point clouds with the radar images. These high-resolution, 2D projected, cropped instances of the map serve as the ground truth images, while the pixel-wise matching radar images serve as the inputs for GAN training. Fig \ref{fig2in} shows the inference process of the GAN, which generates an enhanced version of a radar image solely using the low-resolution radar image input. 

The subsequent sections of the methodology introduce the dataset used in the study, the LiDAR point cloud map generation module, the point cloud map cropping and projection module, the GAN training module, and the evaluation metrics used to assess the performance of the proposed method.

\subsection{Dataset}

This study employed the Boreas dataset \cite{r11} for training and testing the proposed method. The Boreas dataset includes more than 350 km of driving data collected over a year, featuring a sensor platform equipped with cameras, LiDAR, radar, and GNSS. The main reason for using the Boreas dataset is its comprehensive data collection throughout the year under different weather conditions such as sunny, snowy, and rainy weather. This diversity enables effective testing of the proposed method under different weather conditions. The radar and LiDAR sensors used for data capture in the Boreas dataset are a Navtech CIR204-H $360\degree$ radar and a Velodyne Alpha Prime LiDAR. The radar images are available in both 2D polar and Cartesian formats. For this study, the radar images in the Cartesian format were utilized. The data sequences selected for the training and testing phases are specified in Table \ref{table1}. The training of the GAN was done using a sequence from the Boreas dataset captured in good weather conditions. This choice was made because adverse weather conditions can degrade the quality of LiDAR scans, making it challenging to generate high-quality LiDAR point cloud maps. These high-quality maps are essential for obtaining accurate ground truth images to train the GAN. For testing, data sequences from the Boreas dataset representing different weather conditions, including good, snow, and rain were employed.

\begin{table}[h]
\centering
\caption{Details of the Datasets used in Training and Testing}
\begin{tabular}{|c|c|c|c|c|c|}
\hline
\textbf{No.} & \textbf{Boreas sequence}  & \begin{tabular}[c]{@{}c@{}}\textbf{Weather} \\ \textbf{condition}\end{tabular} & \begin{tabular}[c]{@{}c@{}}\textbf{Total} \\ \textbf{images}  \\ \textbf{used}\end{tabular} & \begin{tabular}[c]{@{}c@{}}\textbf{Train} \\ \textbf{images}\end{tabular} & \begin{tabular}[c]{@{}c@{}}\textbf{Test} \\ \textbf{images}\end{tabular} \\ \hline
1   & 2021-09-02-11-42 & Sunny                                                        & 2539                                                             & 2032                                                    & 507                                                    \\ \hline
2   & 2021-01-26-10-59 & Snow                                                         & 984                                                              & -                                                       & 984                                                    \\ \hline
3   & 2021-06-29-18-53 & Rain                                                         & 927                                                              & -                                                       & 927                                                    \\ \hline
\end{tabular}
\label{table1}
\end{table}

\subsection{LiDAR Point Cloud Map Generation}

\begin{figure}[t]
     \centering
     \begin{subfigure}[b]{0.15\textwidth}
         \centering
         \includegraphics[width=\textwidth]{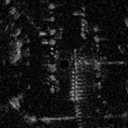}
         \caption{}
         \label{fig3a}
     \end{subfigure}
     \hfill
     \begin{subfigure}[b]{0.15\textwidth}
         \centering
         \includegraphics[width=\textwidth]{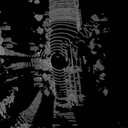}
         \caption{}
         \label{fig3b}
     \end{subfigure}
     \hfill
     \begin{subfigure}[b]{0.15\textwidth}
         \centering
         \includegraphics[width=\textwidth]{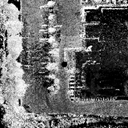}
         \caption{}
         \label{fig3c}
     \end{subfigure}
     \hfill

\caption{Illustration of pixel-wise matching (a) radar image, (b) 2D projected, raw LIDAR scan, (c) 2D projected cropped instance of a LiDAR point cloud map of a sample location from the Boreas dataset.}
\label{fig3}
\end{figure}

For the training of the proposed GAN model, rather than using individual raw LiDAR scans as ground truth images, 2D projected, cropped instances from a LiDAR point cloud map were used. This approach was selected because a cropped instance of a LiDAR point cloud map provides a more comprehensive view of the surroundings, with higher resolution and a broader detection range than a single LiDAR scan \cite{r25}. Figure \ref{fig3} illustrates the pixel-wise matching of a radar image with a 2D projected, a raw LiDAR scan, and a 2D projected cropped instance from a LiDAR point cloud map for a sample location in the Boreas dataset. A comparison between Figures \ref{fig3b} and \ref{fig3c} shows that the cropped instance of the map offers denser information and a greater detection range than a raw LiDAR scan. The methods employed for point cloud map cropping and 2D projection are explained in detail in the following section.

The raw LiDAR scans were accumulated based on their spatial positions and orientations to create a high-resolution point cloud map. Each LiDAR scan was transformed into a global reference frame using its affine transformation matrix, which incorporates both translation and rotation. The affine transformation matrix is composed of translation components ($x, y, z$) and rotation components ($roll, pitch, yaw$) relative to the initial position of the data sequence. These transformation parameters were derived from the GNSS data of the Boreas dataset. After this transformation, the LiDAR scans were accumulated into a single point cloud map. To ensure uniformity and manage data volume, each LiDAR scan was downsampled using a voxel grid filter with a leaf size of $0.8 \times 0.8 \times 0.8$~$\mathrm{m^3}$ before accumulation. This leaf size helps prevent memory overflow during the accumulation process and maintains sufficient point cloud density to distinctly represent objects in the surroundings simultaneously.

\subsection{LiDAR Point Cloud Map Cropping and 2D Projection}\label{next}

The LiDAR point cloud map was cropped into instances as the first step of obtaining pixel-wise matching high-resolution ground truth images and radar image pairs for the training of the GAN. Firstly, the timestamp of each radar image was compared with the timestamps of the poses of the  LiDAR point cloud map. Then the timestamp-wise nearest pose was selected, and a $200 \times 200 \times Z_r$~$\mathrm{m^3}$ box was cropped from the LiDAR point cloud map, centering the nearest pose. $Z_r$ is the range of the $Z$ axis of the LiDAR point cloud map. The dimensions of the bounding box were chosen to ensure that the range of cropped LiDAR point cloud instances matches the detection range of the radar images. These cropped  LiDAR point cloud instances were then transformed into the radar coordinate frame using the LiDAR-to-radar transformation matrix. Finally, these cropped LiDAR point cloud instances were projected onto a 2D plane and transformed into images of $256\times256$ pixel resolution creating high resolution ground truth images. This transformation involved projecting the $x$ and $y$ coordinates of each point onto an image plane while discarding the $z$ coordinate. The intensity of each point was then mapped to the grayscale value of the corresponding pixel in the image. In cases where multiple points corresponded to a single pixel, only the point with the highest $z$ value was considered, and the rest were disregarded.

\subsection{GAN Model and Training}

The proposed method employed the Pix2Pix GAN \cite{n9} for radar image enhancement. The Pix2Pix model uses a conditional GAN, which learns a mapping from an input image  
$x$ and a random noise vector $z$ to an output image $y$ (\(G : \{x, z\} \rightarrow y\)).
In this study, the Pix2Pix model was trained using 2539 pixel-wise matching radar images and 2D projected cropped instances from a high-resolution LiDAR point cloud map. The GAN was trained using a V100 GPU in Compute Canada cloud computers. Training of the GAN involved 1000 epochs with a batch size of 32. Furthermore, augmentation steps were applied during training with a probability of 30\%. These augmentation techniques included random horizontal and vertical shifts, horizontal and vertical flips, zooming, shadows, adjustments in brightness, contrast, and rotation. These steps help the GAN to learn from a diverse range of scenarios, enhancing its ability to generalize across different weather and driving conditions. The training hyperparameters were determined through exhaustive simulation studies. The training time for the GAN was approximately 80 hours, and the average inference time on a V100 GPU is 57.35 milliseconds.

\subsection{Evaluation Metrics}\label{em}

To evaluate the performances of the developed GAN model, peak signal-to-noise ratio (PSNR), structural similarity index (SSIM), and regional mutual information (RMI) were used \cite{r17}.

\section{RESULTS}

The performance evaluation of the proposed GAN model was conducted using three distinct testing datasets from the Boreas dataset, as outlined in Table \ref{table1}. Datasets 1, 2, and 3 were collected in good, snowy, and rainy weather conditions, respectively. Dataset 1  was used for both training and validation of the GAN, whereas datasets 2 and 3 are unseen datasets for the GAN. The performance of the proposed method for dataset 1 was assessed using quantitative metrics introduced in section \ref{em}, with the outcomes detailed in Table \ref{table2}. This table contains two sets of metric values. The first set (i.e., ``Input Radar Image") reports the average PSNR, SSIM, and RMI values, which are derived from comparing low-quality input radar images against high-quality ground truth images. Conversely, the second set (i.e., ``Enhanced Image") reports the average PSNR, SSIM, and RMI values for the GAN-enhanced images relative to the high-quality ground truth images. To the best of our knowledge, no similar studies on radar image enhancement are available in the literature for comparison. Consequently, the average PSNR, SSIM, and RMI values of the input radar images were used as a benchmark to evaluate the GAN-generated images.

\begin{table}[h]
\centering
\caption{Quantitative Results of Dataset 1}
\begin{tabular}{|c|c|c|c|}
\hline
\textbf{Data Type} & \textbf{PSNR} & \textbf{SSIM} & \textbf{RMI} \\
\hline
Input Radar Image  & 28.6377 & 0.0942   & 0.0099      \\
\hline
Enhanced Image & \textbf{30.0533} & \textbf{0.3384}   & \textbf{0.6512}        \\
\hline
\end{tabular}
\label{table2}
\end{table}

As depicted in Table \ref{table2}, the low SSIM and RMI values, along with the comparatively low PSNR value of the input radar images, indicate their poor quality relative to the high-quality ground truth images. Conversely, the higher SSIM and RMI values of the GAN-generated images demonstrate that they are structurally more similar to and exhibit greater mutual alignment with the high-resolution ground truth images. Additionally, the increased PSNR value suggests that the GAN-generated images are of higher quality and more closely resemble the ground truth images.

\begin{figure}[tb]
  \centering
  \includegraphics[width=\columnwidth]{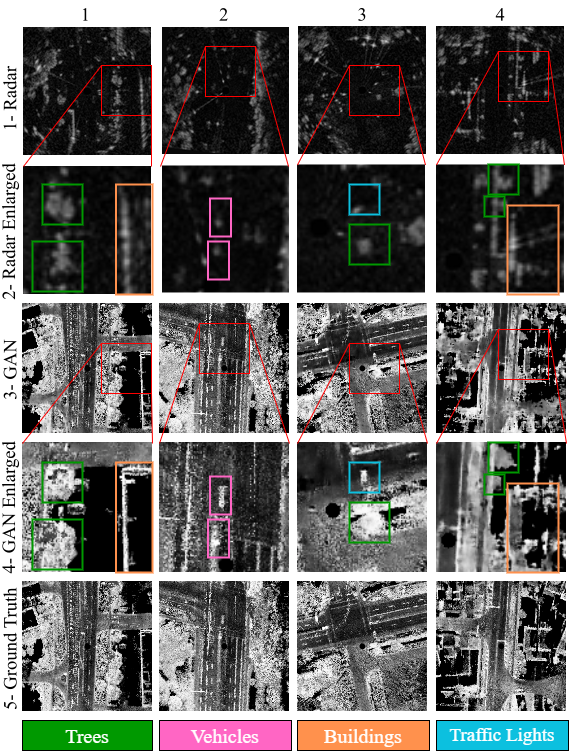}
    \caption{Illustration of sample results of dataset 1, which was collected in good weather conditions. Row 1: input radar images. Row 2: enlarged versions of the red rectangular area of the images in row 1. Row 3: enhanced images generated by the GAN. Row 4: enlarged version of the red rectangular area of the images in row 3. Row 5: corresponding ground truth images. Sample numbers are shown at the top of the figure.}
    \label{fig5}
\end{figure}

A set of sample input radar images, GAN-generated images, and ground truth images from dataset 1 (i.e., sunny weather dataset) are displayed in rows 1, 3, and 5 of Fig. \ref{fig5}, respectively. The images in rows 3 and 5 illustrate that the GAN-generated images are comparable to the ground truth images for dataset 1. This qualitative assessment supports the observations provided by the quantitative results in Table \ref{table2}.

Row 2 and 4 of Fig. \ref{fig5} display enlarged versions of the red rectangular areas of the input radar images in row 1 and GAN-generated enhanced images in row 3, respectively. In the enlargements, some objects are labeled with bounding boxes. These labels are obtained through a careful examination of the location-wise corresponding radar images, LiDAR point clouds, and camera images of dataset 1. These enlargements are used to demonstrate the enhancements in detail and overall quality of the objects in the GAN-generated images relative to the input radar images. By examining the labeled objects in rows 2 and 4 of Fig. \ref{fig5}, it is evident that the GAN has effectively enhanced the details and quality of the objects in the input radar images. For instance, in sample 1, 3, and 4 of Fig. \ref{fig5}, the trees are more distinctly represented with a greater number of pixels in the GAN-generated images compared to the input radar images. In sample result 2, it can be seen that the vehicles that are barely visible in the input radar image are enhanced in the GAN-generated image. Similarly, the traffic light in the sample result 3 is more visible. Additionally, the GAN has improved the representation of features of the buildings, such as walls, and has reconstructed missing information, as seen in sample results 1 and 4.

\begin{figure}[t]
  \centering
  \includegraphics[width=\columnwidth]{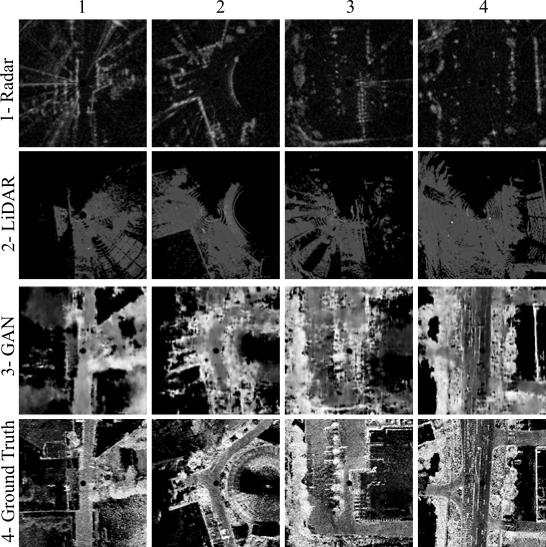}
    \caption{Illustration of sample results of dataset 2, which is collected under snowy weather conditions. Row 1: input radar images. Row 2: corresponding 2D projected, raw LIDAR scans. Row 3: enhanced images generated by the GAN. Row 4: location-wise corresponding ground truth images (cropped from LiDAR point cloud map) from dataset 1. Sample numbers are shown at the top of the figure.}
    \label{fig6}
\end{figure}

\begin{figure}[h]
  \centering
  \includegraphics[width=\columnwidth]{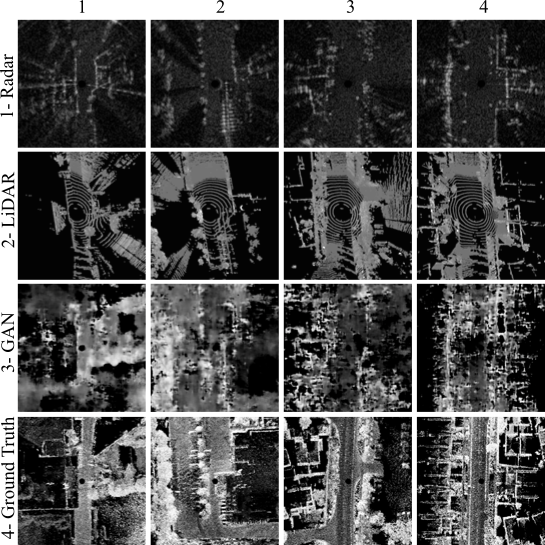}
    \caption{Illustration of sample results of dataset 3, which is collected under rainy weather conditions. Row 1: input radar images. Row 2: corresponding 2D projected, raw LIDAR scans. Row 3: enhanced images generated by the GAN. Row 4: location-wise corresponding ground truth images (cropped from LiDAR point cloud map) from dataset 1. Sample numbers are shown at the top of the figure.}
    \label{fig7}
\end{figure}

A set of sample input radar images, their corresponding 2D projected LiDAR scans, and GAN-generated images from datasets 2 and 3 (i.e., snowy and rainy weather datasets) are displayed in rows 1, 2, and 3 of Fig. \ref{fig6} and \ref{fig7}. Quantitative results were not attainable for datasets 2 and 3 as snow and rain weather conditions have affected the quality of the LiDAR point clouds, making it challenging to generate LiDAR point cloud maps for obtaining ground truth images. However, to qualitatively compare the GAN-generated images of datasets 2 and 3, location-wise corresponding ground truth images from the training dataset were utilized and are shown in the fourth row of Fig \ref{fig6} and \ref{fig7}. It is important to note that these images do not necessarily pixel-wise correspond to those in other rows. This discrepancy arises because the data were collected at different times, potentially leading to the presence or absence of dynamic objects and variations in the driving path, among other factors. 

When comparing the radar images and individual LiDAR point clouds depicted in the first and second rows of Fig \ref{fig6}, it is evident that the LiDAR point clouds were affected by snowy weather conditions, resulting in numerous missing data points and a notable reduction in the detection range. However, upon comparing the images generated by the GAN with the ground truth images in the third and fourth rows, it is evident that the GAN model has successfully enhanced the input radar images and reconstructed the missing information. Furthermore, the enhanced images closely resemble the ground truth images, showcasing the effectiveness of the proposed method, even under snowy weather conditions.

The results for dataset 3 (i.e., rainy weather dataset) shown in Fig \ref{fig7} indicate that the GAN-generated images resemble the ground truth images to some extent but not as accurately as those in sunny and snowy weather datasets. This discrepancy could potentially be due to the influence of rainy weather conditions on the radar images. It is important to note that this study exclusively utilized data collected in sunny weather conditions for the training of the GAN. This finding highlights the need for more extensive training of the GAN, incorporating data augmentation to simulate the effects of rain on radar images, enabling the GAN to learn and generalize these conditions better.


The results from different weather conditions demonstrate that the GAN-based method developed in this study can enhance radar images, relying solely on the radar data itself, to provide a more useful representation of surrounding objects of AVs. These enhanced images closely resemble the ground truth images, even in adverse weather conditions. These results provide a promising direction for continued research in GAN-based radar image enhancement.

\section{CONCLUSIONS}

This paper introduced a GAN-based method for radar image enhancement. The proposed method exclusively utilizes radar sensors to produce enhanced versions of radar images. These enhanced images more accurately depict the details and features of the objects in the environment compared to the radar images, thereby facilitating improved object identification. Based on both qualitative and quantitative results, it can be stated the proposed method performs satisfactorily well in good weather conditions. Furthermore, the qualitative results indicate that the proposed method is robust for snowy weather conditions and generates comparable images to the ground truth images. These results indicate the effectiveness of the proposed method when the camera and LiDAR sensors are unavailable or when these sensors are compromised by adverse weather conditions. In rainy weather conditions, the proposed method experiences a slight performance drop compared to other weather conditions. This suggests that further improvements and more generalized training are needed for the GAN model. Nonetheless, These findings present a promising direction for further research in GAN-based methods for LiDAR-assisted radar image enhancement. As future work, it is expected to develop a tailor-made GAN for radar image enhancement and semantic segmentation. Furthermore, this customized GAN will be trained using data gathered from various locations and various radar sensors supplemented with weather-simulated augmentation steps to improve its generalization.

\addtolength{\textheight}{-12cm}   





\end{document}